\documentclass[letterpaper, 10 pt, journal, twoside]{IEEEtran}

 \usepackage[utf8]{inputenc}

\usepackage[exponent-product=.,detect-weight]{siunitx}
\usepackage[hyphens]{url}
\usepackage[breaklinks]{hyperref}
\hypersetup{hidelinks}
\usepackage{amsmath, 
	amsfonts, 
	amssymb, 
	bm,
	mathtools,
	amsthm,
	siunitx}
\usepackage{graphicx}
\usepackage[dvipsnames]{
		xcolor
	}
\usepackage{booktabs}
\usepackage{multirow}
\usepackage{tikz,
	pgfplots,
	pgfplotstable,
	eso-pic}
\usetikzlibrary{patterns,
	plotmarks}
\pgfplotsset{compat=1.12}

\usepackage[english,
	ruled,
	]{algorithm2e}
	
\usepackage[nospace]{cite}

\usepackage{blindtext,
	ifpdf}
\usepackage{setspace}
\usepackage{paralist}
\usepackage{filecontents}
\usepackage{ifthen}

\usetikzlibrary{shapes,arrows,shadows,snakes,automata,backgrounds,petri}
\usetikzlibrary{pgfplots.groupplots}
\usepgfplotslibrary{statistics}
\pgfplotsset{grid style={dashed,gray}}
\pgfplotsset{major grid style={dashed,gray}}
\pgfplotsset{minor grid style={dotted,gray!50}}

\usepackage{afterpage}
\usepackage{,times}
\usepackage[top=54pt, bottom=54pt, left=54pt, right=54pt]{geometry}

\newcommand{\bfa}{\mathbf{a}}

\newcommand{\bfA}{\mathbf{A}}
\newcommand{\bfb}{\mathbf{b}}
\newcommand{\hbfb}{\hat{\mathbf{b}}}

\newcommand{\bfc}{\mathbf{c}}
\newcommand{\bfC}{\mathbf{C}}

\newcommand{\bfg}{\mathbf{g}}

\newcommand{\bfI}{\mathbf{I}}

\newcommand{\calL}{\mathcal{L}}

\newcommand{\bfM}{\mathbf{M}}

\newcommand{\bfR}{\mathbf{R}}

\newcommand{\hbfR}{\hat{\bfR}}
\newcommand{\bbR}{\mathbb{R}}

\newcommand{\bfS}{\mathbf{S}}

\newcommand{\bfu}{\mathbf{u}}

\newcommand{\bfv}{\mathbf{v}}

\newcommand{\boeta}{\boldsymbol{\eta}}

\newcommand{\boomega}{\boldsymbol{\omega}}

\newcommand{\hboomega}{\hat{\boomega}}

\newcommand{\bfzero}{\mathbf{0}}

\newtheoremstyle{mystyle}
{}
{}
{\itshape}
{}
{\bfseries}
{.}
{ }
{}

\newboolean{silvere}
\setboolean{silvere}{true}

\ifthenelse{\boolean{silvere}}{}{
\usepgfplotslibrary{external}
\tikzexternalize
}

\begin{document}
\markboth{IEEE Robotics and Automation Letters. Preprint Version. Accepted
June, 2020}
{Brossard \MakeLowercase{\textit{et al.}}: Denoising IMU Gyroscopes with Deep Learning for Open-Loop Attitude Estimation}
\title{\vspace*{0.0cm} Denoising IMU Gyroscopes with Deep Learning for Open-Loop Attitude Estimation}

\author{Martin Brossard$^{1}$, Silv\`ere Bonnabel$^{1,2}$, and Axel Barrau$^{1,3}$%
\thanks{Manuscript received: February 24, 2020; Revised April, 20, 2020;
Accepted June, 8, 2020.}
\thanks{This paper was recommended for publication by Editor Sven Behnke upon
evaluation of the Associate Editor and Reviewers' comments.}
\thanks{$^{1}$Martin Brossard, Silv\`ere Bonnabel and Axel Barrau are with Centre for Robotics, MINES ParisTech, PSL Research University, Paris, France {\tt\small martin.brossard@mines-paristech.fr}}%
\thanks{$^{2} $Silv\`ere Bonnabel is with University of New Caledonia, ISEA, Noumea Cedex, New Caledonia {\tt\small silvere.bonnabel@unc.nc} }%
\thanks{$^{3} $Axel Barrau is with Safran Tech, Groupe Safran, Magny Les Hameaux Cedex, France {\tt\small axel.barrau@safrangroup.com} }%
\thanks{Digital Object Identifier (DOI): see top of this page.}
}

\maketitle

\begin{abstract}
This paper proposes a learning method for denoising gyroscopes  of Inertial Measurement
Units (IMUs) using ground truth data, and estimating in real time the orientation (attitude) of a robot in dead reckoning. The obtained algorithm outperforms the state-of-the-art on the
(unseen) test sequences.
The obtained performances are achieved thanks to a well-chosen model,  a proper loss   function for orientation
increments, and through the identification of   key
points when training with high-frequency inertial data. Our approach builds upon a neural network based on dilated convolutions, without requiring any recurrent
neural network. We demonstrate how efficient our strategy is for 3D attitude
estimation on the EuRoC and TUM-VI datasets.  Interestingly, we observe our dead reckoning algorithm manages     to beat  top-ranked
visual-inertial odometry systems in terms of attitude estimation although it
does not use vision sensors.   We believe this paper offers new
perspectives for visual-inertial localization and constitutes a step toward more efficient
learning methods involving IMUs. Our open-source implementation is available  at \texttt{\url{https://github.com/mbrossar/denoise-imu-gyro}}.
\end{abstract}

\begin{IEEEkeywords}
	localization,  deep  learning  in  robotics and automation, autonomous systems navigation
\end{IEEEkeywords}

\section{Introduction}\label{sec:int}

\IEEEPARstart{I}{Inertial} Measurement Units (IMUs) consist of gyroscopes that measure  angular velocities, i.e. the rate of change of the sensor’s orientation, and accelerometers that
measure proper accelerations
\cite{kokUsing2017}. IMUs allow estimating a robot's trajectory relative to its starting position, a task called odometry \cite{scaramuzzaVisualInertial2019}.

Small and cheap IMUs are ubiquitous in smartphones, industrial and robotics applications, but suffer from
difficulties to estimate sources of error, such as axis-misalignment, scale factors and time-varying offsets
\cite{rehderExtending2016, rohacCalibration2015}. Hence, IMU signals are not only noisy, but they are biased. In the
present paper, we propose to leverage deep learning for denoising the gyroscopes
(gyros) of an IMU, that is, reduce noise and biases.
As a byproduct, we obtain accurate attitude (i.e. orientation) estimates simply by open-loop integration of the obtained noise-free gyro measurements.

\subsection{Links and Differences with Existing Literature}

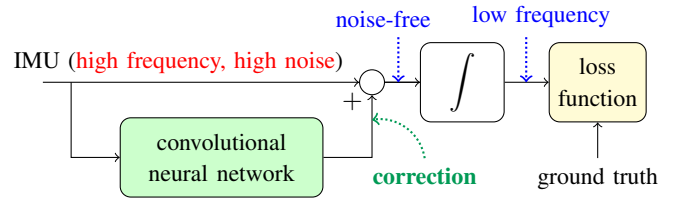
\begin{figure}
\centering
\ifthenelse{\boolean{silvere}}{
	\includegraphics{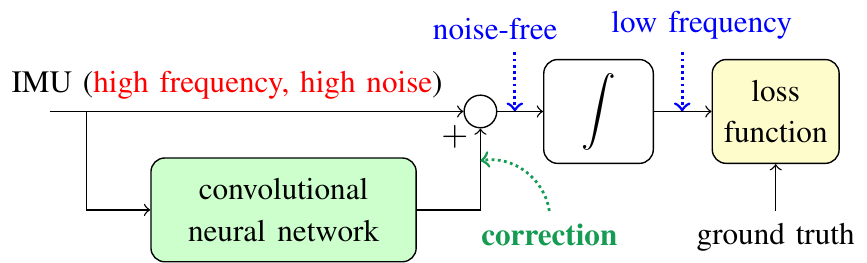}
}{
	\tikzstyle{block} = [draw, rectangle, minimum height=3em, minimum
width=0.5em, text width=2.5em, text centered, rounded corners]
\tikzstyle{sum} = [draw, circle, node distance=1cm]
\tikzstyle{pinstyle} = [pin edge={to-,thin,black}]
\begin{tikzpicture}[auto, node distance=1cm]
    \node (input) at (0, 0) {};
    \node [sum] (sum) at (4.5, 0) {};
	\node [block, text width=7em, fill=green!20] (cnn) at (2.5, -1) {\small convolutional neural network};
	\node [block, right of=sum, node distance=1.2cm] (integrate) {$\displaystyle\int$};
\node [block, right of=integrate, pin={[pinstyle]below:\small ground truth},  node distance=1.8cm, text width=3em,fill=yellow!20] (loss) {\small loss function};

\draw [draw,->] (input) -- node [xshift=-0.3cm] {\small IMU (\textcolor{red}{high frequency, high noise})}
(sum);
	\draw [draw,->] (sum) -- (integrate) -- (loss);
	\draw [draw,->] (sum) -- (integrate);
	\draw [->] (0.5, 0) |- (cnn);
	\draw [->] (cnn) -| node[pos=0.95] {$+$} (sum);

	\draw[->, blue, densely dotted, thick] (4.85,0.6) node[above,xshift=-0.2cm] {\small noise-free} --
(4.85,0);
\draw[->, blue, densely dotted, thick] (6.55,0.6) node[above,xshift=0.2cm] {\small low frequency} --
(6.55,0);

\draw[->, densely dotted, thick, ForestGreen] (5.2,-1.01) node[below,xshift=0.0cm] {\small \textbf{correction}} to[bend right=45] (4.5,-0.5);

\end{tikzpicture}
}
\vspace*{-0.5cm}
\caption{Schematic illustration of the proposed method. The convolutional neural network computes gyro corrections (based on past IMU measurements) that filter undesirable errors in the raw IMU signals. We then perform \emph{open-loop} time integration on the noise-free measurements for regressing low frequency errors between ground truth and estimated orientation increments.\label{fig:schema}\vspace*{-0.5cm}}
\end{figure}

IMUs are generally coupled with complementary sensors to obtain robust pose
estimates in sensor-fusion systems \cite{forsterOnManifold2017}, where the supplementary information is provided 
by either cameras in Visual-Inertial Odometry (VIO)
\cite{genevaOpenVINS2019,qinVINSMono2018,svachaInertial2019}, LiDAR,   GNSS, or may step from side information about the model 
\cite{brossardRINSW2019,brossardAIIMU2019,madgwickEstimation2011,solinInertial2018}. To obtain accurate pose estimates, a
proper IMU calibration is required, see e.g. the widely used \emph{Kalibr} library
\cite{furgaleUnified2013,rehderExtending2016}, which computes offline the underlying IMU intrinsic
calibration parameters, and extrinsic parameters between the camera and IMU. Our approach, which is recapped in Figure
\ref{fig:schema}, is applicable to any system equipped
with an IMU. It estimates offline the IMU calibration parameters and extends methods such as \cite{furgaleUnified2013, rehderExtending2016}
to time-varying signal corrections.

Machine learning (more specifically  deep learning) has been recently leveraged to perform LiDAR,
visual-inertial, and purely inertial
localization, where methods are divided  into supervised \cite{chenIONet2018,clarkVINet2017,yanRIDI2018,esfahaniOriNet2020} and unsupervised \cite{almaliogluSelfVIO2019}.
Most works extract relevant features in the sensors' signals which are propagated in time through
recurrent neural networks, whereas \cite{yanRoNIN2019} proposes convolutional
neural networks for pedestrian inertial navigation. A related approach  \cite{nobreLearning2019}
applies reinforcement learning for
guiding the user to properly calibrate visual-inertial rigs.
Our method is supervised (we require ground truth poses for training), leverages convolutions rather than recurrent architectures, and  
outperforms the latter approach. We obtain excellent results while requiring considerably fewer
data and less time. Finally, the reference
\cite{esfahaniOriNet2020}  estimates orientation with an IMU and recurrent neural networks, but our approach proves simpler.  

\subsection{Contributions}

Our main contributions are as follows:
\begin{itemize}
\item detailed modelling  of the problem of learning orientation increments from low-cost IMUs;
\item the convolutional neural network which regresses gyro corrections and whose features are carefully selected;
\item the training procedure which involves a trackable loss function for estimating relative orientation increments;
\item the approach evaluation  on datasets acquired by a drone and a hand-held device \cite{burriEuRoC2016,schubertTUM2018}, where our method outperforms \cite{esfahaniOriNet2020} and competes with VIO methods \cite{genevaOpenVINS2019,qinVINSMono2018} although it does not use vision;
\item perspectives towards more efficient VIO and IMU based learning methods;
\item a publicly available open-sourced code, where training is done in 5 minutes per dataset.
\end{itemize}

\section{Kinematic \& Low-Cost IMU Models}\label{sec:model}

We detail in this section our model.

\subsection{Kinematic Model based on Orientation Increments}

The 3D orientation of a rigid platform is obtained by integrating orientation increments, that is, gyro
outputs of an IMU, through
\begin{align}
	\bfR_{n} = \bfR_{n-1}\exp\left(\boomega_n dt\right), \label{eq:ori}
\end{align}
where the rotation matrix $\bfR_n \in SO(3)$ at timestamp $n$ maps the IMU frame to the global frame, the angular velocity $\boomega_n \in \bbR^3$ is averaged during  $dt$, and with $\exp(\cdot)$ the $SO(3)$ exponential map. The model \eqref{eq:ori} successively integrates in open-loop  $\boomega_n$ and \emph{propagates} estimation errors. Indeed, let $\hbfR_{n}$ denotes an estimate of $\bfR_{n}$. The error present  in $\hbfR_{n-1}$ is propagated in $\hbfR_n$ through  \eqref{eq:ori}.

\subsection{Low-Cost Inertial Measurement Unit (IMU) Sensor Model}
The IMU provides noisy and biased measurements of angular rate $\boomega_n$ and
specific acceleration $\bfa_n$ at high frequency (\SI{200}{Hz} in our experiments) as, see 
\cite{rohacCalibration2015, rehderExtending2016},
\begin{align}
 \bfu_n^{\mathrm{\textsc{imu}}}= \begin{bmatrix}
\boomega_n^{\mathrm{\textsc{imu}}}\\
\bfa_n^{\mathrm{\textsc{imu}}}
\end{bmatrix} &= \bfC \begin{bmatrix}\boomega_n \\ \bfa_n \end{bmatrix} + \bfb_{n} + \boeta_{n}, \label{eq:imu}
\end{align}
where $\bfb_{n} \in \bbR^6$ are quasi-constant biases, $\boeta_{n} \in \bbR^6$ are
commonly assumed zero-mean white Gaussian noises,  and
\begin{align} \bfa_n=
	\bfR_{n-1}^T\left(\left(\bfv_{n}-\bfv_{n-1}\right)/dt-\bfg\right) \in \bbR^3 \label{eq:acc}
\end{align}
is the acceleration in the IMU frame without the effects of gravity $\bfg$, with
$\bfv_n
\in \bbR^3$ the IMU velocity expressed in the global frame. The intrinsic \emph{calibration} matrix 
\begin{align}
\bfC = \begin{bmatrix}  \bfS_{\boomega}\bfM_{\boomega} & \bfA \\ \bfzero_{3\times3}  &\bfS_{\bfa} \bfM_{\bfa}
\end{bmatrix}\approx \bfI_6 \label{eq:calib}
\end{align} 
contains the information for correcting signals: axis misalignments (matrices $\bfM_{\boomega}\approx \bfI_3$ and $\bfM_{\bfa}\approx \bfI_3$); scale factors (diagonal matrices $\bfS_{\boomega}\approx \bfI_3$ and $\bfS_{\bfa}\approx \bfI_3$); and linear accelerations on
gyro measurements, a.k.a. g-sensitivity (matrix $\bfA\approx \bfzero_{3\times3}$). Remaining intrinsic
parameters, e.g. level-arm between gyro and accelerometer, can be found in
\cite{rohacCalibration2015, rehderExtending2016}.

We now make three remarks regarding \eqref{eq:ori}-\eqref{eq:calib}:
\begin{enumerate}
	\item equations \eqref{eq:imu}-\eqref{eq:calib} represent a model that
	\emph{approximates} reality. Indeed, calibration parameters $\bfC$ and biases
	$\bfb_n$ should both depend on time as they vary
	 with temperature and stress \cite{kokUsing2017, rohacCalibration2015}, 
	 but are difficult to
	estimate in real-time. Then, vibrations and platform excitations due to,  e.g.
	rotors, make  Gaussian noise $\boeta_n$ colored in
	practice \cite{luIMUBased2019}, albeit commonly assumed white;
	\item substituting actual measurements 
	$\boomega_{n}^{\mathrm{\textsc{imu}}}$ in place of true value $\boomega_{n}$
	in \eqref{eq:ori} leads generally to quick drift (in a few seconds) and poor orientation estimates;
	\item in terms of method evaluation, one should
always compare the learning method with respect to results obtained with a properly calibrated IMU
	as a proper estimation of the
	parameters $\bfC$ and $\bfb_n$ in \eqref{eq:imu} actually leads to surprisingly precise results, see
	Section
	\ref{sec:experiments}.
\end{enumerate}

\section{Learning Method for Denoising the IMU}\label{sec:method}

We describe in this section our approach for regression of noise-free gyro increments
$\hboomega_n$ in \eqref{eq:imu} in order to  obtain accurate orientation estimates by integration of $\hboomega_n$ in \eqref{eq:ori}. Our
goal thus boils down to estimating $\bfb_n$, $\boeta_n$, and correcting the misknown
$\bfC$.

\subsection{Proposed Gyro Correction Model}
Leveraging the analysis of Section \ref{sec:model}, we compute the noise-free increments as
\begin{align}
	\hboomega_n = \hat{\bfC}_{\boomega} \boomega_n^{\mathrm{\textsc{imu}}} + \tilde{\boomega}_n, \label{eq:homega}
\end{align}
with $\hat{\bfC}_{\boomega} = \hat{\bfS}_{\boomega} \hat{\bfM}_{\boomega} \in
\bbR^{3\times3}$ the intrinsic parameters that account for gyro axis-misalignment and
scale factors, and where the gyro bias is included in the gyro correction
$\tilde{\boomega}_n = \hat{\bfc}_n + \hbfb$,
where $\hat{\bfc}_n$ is  time-varying and $\hbfb$ is the static bias. Explicitly considering the small accelerometer influence  $\bfA$, see
\eqref{eq:imu}-\eqref{eq:calib}, does not affect the results so it is ignored.

\begin{figure}
	\centering
\ifthenelse{\boolean{silvere}}{
	\includegraphics{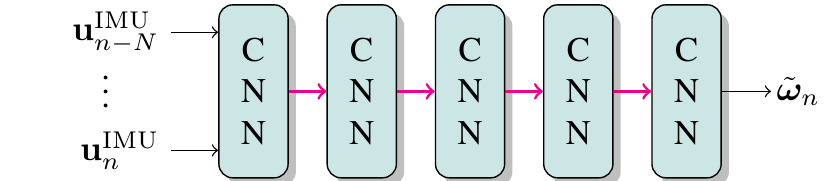}
	~
}{
	\begin{tikzpicture}
\node (cnn0) [draw, text width=8em, rounded corners, minimum height=5em, minimum width=2em, drop shadow, text centered,fill=teal!20, text width=1em]  {C N N};

\path (cnn0)++(1.1,0) node (cnn1)[draw, text width=8em,
minimum height=5em, minimum width=2em, rounded corners, drop shadow, text centered,fill=teal!20,
text width=1em]  {C N N};
\path (cnn1)++(1.1,0) node (cnn2)[draw, text width=8em,
minimum height=5em, minimum width=2em, rounded corners, drop shadow, text centered,fill=teal!20,
text width=1em]  {C N N};
\path (cnn2)++(1.1,0) node (cnn3)[draw, text width=8em,
minimum height=5em, minimum width=2em, rounded corners, drop shadow, text centered,fill=teal!20,
text width=1em]  {C N N};
\path (cnn3)++(1.1,0) node (cnn4)[draw, text width=8em,
minimum height=5em, minimum width=2em, rounded corners, drop shadow, text centered,fill=teal!20,
text width=1em]  {C N N};

\path (cnn0.west)++(-1.1,0.6) node[text width=1cm,align=right] (omega0)
{$\bfu^{\mathrm{IMU}}_{n-N}$};
\path (cnn0.west)++(-1.6,0.1) node[text width=1cm,align=right] (omega2)
{$\vdots$};
\path [draw, ->] (omega0.east) -- ($(cnn0.west)+(0,0.6)$);
\path (cnn0.west)++(-1.1,-0.6) node[text width=1cm,align=right] (omega1)
{$\bfu^{\mathrm{IMU}}_{n}$};
\path [draw, ->] (omega1.east) -- ($(cnn0.west)+(0,-0.6)$);
\path [draw,magenta, thick, ->] (cnn0.east) -- (cnn1.west);
\path [draw,magenta, thick, ->] (cnn1.east) -- (cnn2.west);
\path [draw,magenta, thick, ->] (cnn2.east) -- (cnn3.west);
\path [draw,magenta, thick, ->] (cnn3.east) -- (cnn4.west);
\path [draw, ->] (cnn4.east) -- ++(0.5, 0) node[text width=1cm,align=right] (y) {$\tilde{\boomega}_n$};

\path (0,-0.0) node{};
\end{tikzpicture}
}
\vspace*{-0.2cm}
\small
\newcolumntype{C}[1]{>{\centering\arraybackslash}p{#1}}

\begin{tabular}{c||C{0.57cm}|C{0.57cm}|C{0.57cm}|C{0.57cm}|C{0.57cm}}
		\toprule
		CNN layer \# &1 &2 &3 &4&5\\ \midrule
        kernel dim.  & 7  &  7 & 7  &  7 & 1   \\ \midrule
		dilatation gap   & 1  &  4 & 16  & 64 & 1   \\ \midrule
        channel dim.  & 16  & 32 & 64  & 128 & 1   \\
    \bottomrule
\end{tabular}
\caption{\label{fig:cnn} Proposed neural network structure which computes gyro
correction $\tilde{\boomega}_n$ in \eqref{eq:homega} from the  $N$ past IMU measurements. We set the
Convolutional Neural Networks (CNNs) as indicated in the table,
and define between two convolutional layers a batchnorm layer
\cite{ioffeBatch2015} and a smooth GELU activation function
\cite{ramachandranSearching2018}  (magenta arrows).  The adopted structure defines the window
size $N=\max \left(\text{kernel dim.}\times \text{dilation gap}\right) = 7
\times 64=448$, corresponding to \SI{2.24}{s} of past information, and has been
found after trial-and-error on datasets \cite{burriEuRoC2016,schubertTUM2018}.\vspace*{-0.5cm}}
\end{figure}
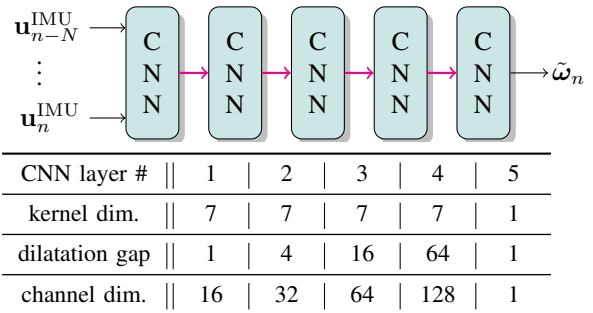

We now search to compute $\tilde{\boomega}_n$ and $\hat{\bfC}_{\boomega}$. The neural
network described in Section \ref{sec:cnn} computes $\tilde{\boomega}_n$ by
leveraging information present in a past local window of size $N$ around
$\boomega_n^{\mathrm{\textsc{imu}}}$. In contrast, we let $\hat{\bfC}_{\boomega}$ be static parameters initialized at
$\bfI_3$ and optimized during training since each considered dataset
uses \emph{one} IMU. The learning problem involving a time varying $\hat{\bfC}_{\boomega}$ and/or multiple
IMUs is let for future works. 

The consequences of opting for the model  \eqref{eq:homega} and the proposed network
structure are as follows. First, the corrected gyro is initialized on  the
	original gyro $\hboomega_n \approx
	\boomega_n^{\mathrm{\textsc{imu}}}$, with $\hat{\bfC}_{\boomega}=\bfI_3$ and
	$\tilde{\boomega}_n \approx \bfzero_3$ before training. This way,   the method  
	improves the estimates as soon as the first training epoch. Then, our method is intrinsically
robust to
overfitting as measurements outside the local window, i.e. whose timestamps are
less than
$n-N$ or greater than $n$, see Figure \ref{fig:cnn}, do not participate in 
infering $\tilde{\boomega}_n$. This allows to train the method with 8 or less minutes  of data, see Section
\ref{sec:dataset}.

\subsection{Dilated Convolutional Neural Network Structure}\label{sec:cnn}

We define here the neural network structure which infers the  gyro correction as
\begin{align}
	\tilde{\boomega}_n = f(\bfu_{n-N}^{\mathrm{\textsc{imu}}},\ldots,\bfu_n^{\mathrm{\textsc{imu}}}), \label{eq:tilde}
\end{align}
where $f(\cdot)$ is the function defined by the neural network.
 The network should
extract information at temporal multi-scales and compute smooth corrections. Note that, the input of the network consists of IMU data, that is, gyros naturally, but also  accelerometers signals. Indeed, from \eqref{eq:acc}, if the
velocity varies slowly between successive increments we have
\begin{align}
	\bfa_{n+1}-\bfa_{n}&\approx -(\bfR_n-
\bfR_{n-1})^T\bfg \nonumber\\&
\approx  -(\exp\left(-\boomega_n dt\right)-\bfI_3) \bfR_{n-1}^T\bfg, \label{eq:new}
\end{align}
which also provides information about angular velocity. This illustrates how $\bfa_{n}$ can be used, albeit the neural network does not assume small velocity variation.

\begin{figure}
	\centering
	\ifthenelse{\boolean{silvere}}{
		\includegraphics{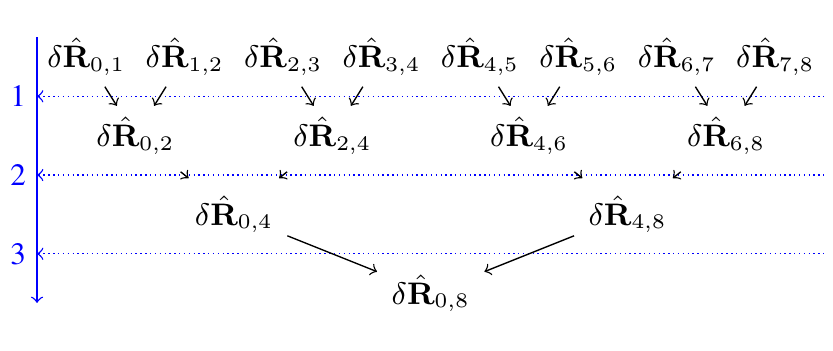}
		~
	}{
		\begin{tikzpicture}
	\small
	\node[circle](01) at (0,0) {$\delta \hbfR_{0,1}$};
	\node[circle](12) at (1,0) {$\delta \hbfR_{1,2}$};
	\node[circle](23) at (2,0) {$\delta \hbfR_{2,3}$};
	\node[circle](34) at (3,0) {$\delta \hbfR_{3,4}$};
	\node[circle](45) at (4,0) {$\delta \hbfR_{4,5}$};
	\node[circle](56) at (5,0) {$\delta \hbfR_{5,6}$};
	\node[circle](67) at (6,0) {$\delta \hbfR_{6,7}$};
	\node[circle](78) at (7,0) {$\delta \hbfR_{7,8}$};

	\node[circle](02) at (0.5,-0.8) {$\delta \hbfR_{0,2}$};
	\node[circle](24) at (2.5,-0.8) {$\delta \hbfR_{2,4}$};
	\node[circle](46) at (4.5,-0.8) {$\delta \hbfR_{4,6}$};
	\node[circle](68) at (6.5,-0.8) {$\delta \hbfR_{6,8}$};

	\node[circle](04) at (1.5,-1.6) {$\delta \hbfR_{0,4}$};
	\node[circle](48) at (5.5,-1.6) {$\delta \hbfR_{4,8}$};

	\node[circle](08) at (3.5,-2.4) {$\delta \hbfR_{0,8}$};

	\draw[blue, densely dotted,<-] (-0.5, -0.4) -- (7.5,-0.4);
	\draw[blue, densely dotted,<-] (-0.5, -1.2) -- (7.5,-1.2);
	\draw[blue, densely dotted,<-] (-0.5, -2) -- (7.5,-2);

	\draw[<-] (01) -- (02);
	\draw[<-] (12) -- (02);
	\draw[<-] (23) -- (24);
	\draw[<-] (34) -- (24);
	\draw[<-] (45) -- (46);
	\draw[<-] (56) -- (46);
	\draw[<-] (67) -- (68);
	\draw[<-] (78) -- (68);

	\draw[->] (02) -- (04);
	\draw[->] (24) -- (04);
	\draw[->] (46) -- (48);
	\draw[->] (68) -- (48);

	\draw[->] (04) -- (08);
	\draw[->] (48) -- (08);

\draw[->,blue] (-0.5,0.2) -- (-0.5,-2.5);
	
\draw[blue] (-0.5,-0.4) node[left] {1};
\draw[blue] (-0.5,-1.2) node[left] {2};
\draw[blue] (-0.5,-2) node[left] {3};
\end{tikzpicture}
	}
	\vspace*{-1.1cm}
\caption{Time efficient computation of the loss \eqref{eq:loss} by viewing the
orientation
	integration \eqref{eq:loss_incr} as a
tree of matrix multiplications. Computation for length $j$ requires $\log_2(j)$
``batch'' 
	operations as denoted by the blue vertical arrow  on the left. We see we need 3 batches of parallel operations for $j=8$ on the chart above. In the same way, we  only need  5 operations
	for $j=32$.\label{fig:tree}\vspace*{-0.5cm}}
	\end{figure}
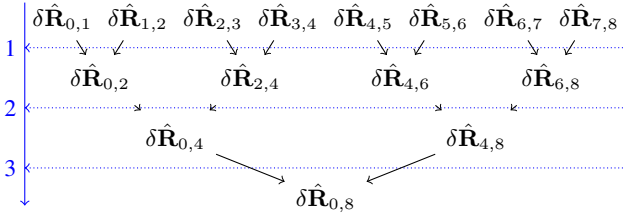

We leverage  dilated convolutions  that infer a correction based on a local
window of $N=448$ previous measurements, which represents \SI{2.24}{s} of information before timestamp $n$ in our experiments.
Dilated convolutions are a method based on  convolutions applied to input with defined dilatation
gap, see   \cite{yuMultiScale2016}, which: $i$) supports exponential expansion of the receptive field, i.e., $N$,
without loss of resolution or coverage; $ii$) is computationally efficient  with few memory
consumption; and $iii$) maintains the temporal ordering of data. We thus expect
the network to detect  and correct various features such as   rotor vibrations that are not modeled in 
\eqref{eq:imu}. Our
configuration given in Figure \ref{fig:cnn} requires learning \num{77052}
parameters, which is \emph{extremely} low and contrasts with recent (visual-)inertial learning methods, see
e.g. \cite{almaliogluSelfVIO2019} Figure 2, where IMU processing only requires more than
\num{2600000} parameters.

	\begin{figure}
		\centering
		\ifthenelse{\boolean{silvere}}{
			\includegraphics{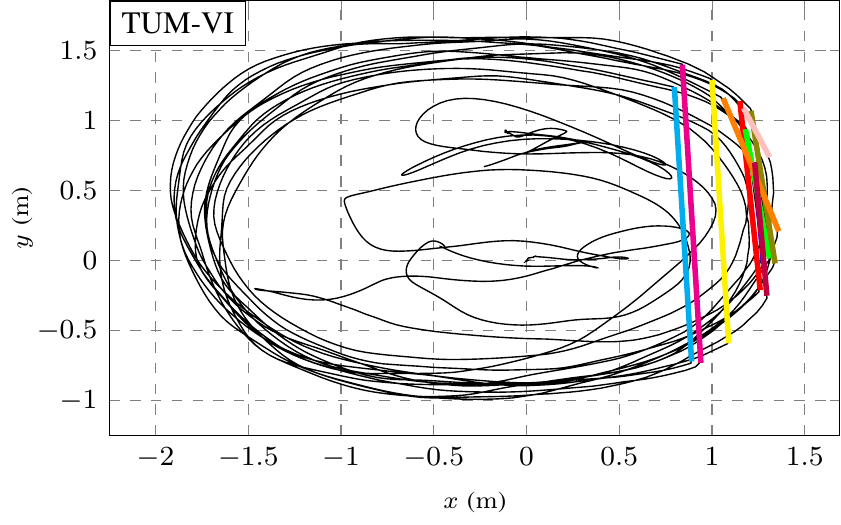}
		}{
			\begin{tikzpicture}
    \begin{axis}[
        footnotesize,
        width=9cm,
        height=6cm,
        xlabel={$x$ (\si{m})},
        ylabel={$y$ (\si{m})},
        xlabel style={font=\scriptsize},
        ylabel style={font=\scriptsize},
        xtick style={font=\scriptsize},
        ytick style={font=\scriptsize},
        xtick={-2,-1.5,...,1.5},
        ylabel shift = -0.2cm,
        ticks=both,
        grid=both,
        title={TUM-VI},font=\small,every axis title/.style={below
right,at={(0,1)},draw=black,fill=white},
    ]

    \addplot[draw=black]  table[x=x,y=y] {figures/2.txt};
    \addplot[draw=cyan, ultra thick] coordinates {(0.89,-0.72) (0.797,1.24)};
    \addplot[draw=magenta, ultra thick] coordinates {(0.94,-0.73) (0.84,1.4)};
    \addplot[draw=yellow, ultra thick] coordinates {(1.09,-0.59) (1,1.29)};
    \addplot[draw=red, ultra thick] coordinates {(1.26,-0.21) (1.15,1.14)};
    \addplot[draw=blue, ultra thick] coordinates {(1.277,-0.02) (1.23,0.705)};
    \addplot[draw=green, ultra thick] coordinates {(1.31,0.015) (1.18,0.94)};
    \addplot[draw=olive, ultra thick] coordinates {(1.34,-0.02) (1.21,1.07)};
    \addplot[draw=orange, ultra thick] coordinates {(1.36,0.21) (1.06,1.16)};
    \addplot[draw=pink, ultra thick] coordinates {(1.31,0.74) (1.17,1.09)};
    \addplot[draw=purple, ultra thick] coordinates {(1.296,-0.25) (1.23,0.70)};
    
    \end{axis}
\end{tikzpicture}
		}
		\vspace*{-0.9cm}
		\caption{Horizontal ground truth trajectory for the sequence \texttt{room} \texttt{1} of \cite{schubertTUM2018}. Ground truth is periodically absent due to occlusions of the hand-held device from the motion capture system, see the color lines on the right of the figure.\label{fig:gt2}\vspace*{-0.5cm}}
	\end{figure}
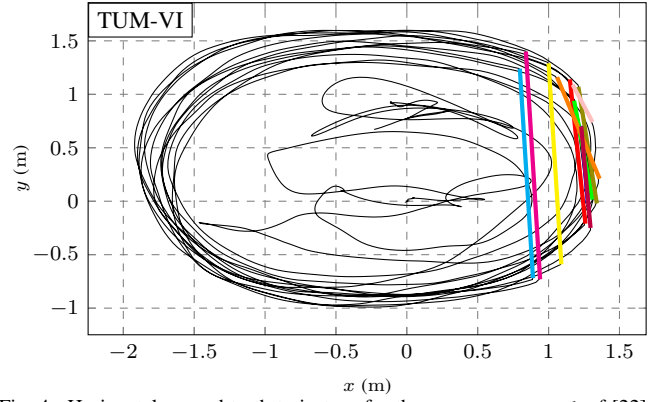

\subsection{Loss Function based on Integrated Gyro Increments}
Defining a loss function directly based on errors
$\boomega_n-\hboomega_n$ requires having a ground truth $\boomega_n$ at IMU
frequency (\SI{200}{Hz}), which is not feasible    in practice as the best 
tracking systems are accurate at \num{20}-\SI{120}{Hz}.  In place, we suggest  defining a loss based on 
the following integrated increments
\begin{align}
	\delta \bfR_{i,i+j} &= \bfR_i^T \bfR_{i+j} = \prod_{k=i}^{i+j-1} \exp\left(\boomega_{k}\right), \label{eq:loss_incr}
\end{align}
i.e., where the IMU frequency is reduced by a factor $j$. We then
compute the loss for a given $j$ as
\begin{align}
	\calL_j &= \sum_{i} \rho\left(\log\left(\delta \bfR_{i,i+j} \delta \hat{\bfR}_{i,i+j}^T \right)\right), \label{eq:loss}
\end{align}
where $\log(\cdot)$ is the $SO(3)$ logarithm map, and  $\rho(\cdot)$ is the
Huber loss function. We set in our experiments the Huber loss parameter to
0.005, and   define our loss function as
\begin{align}
	\calL = \calL_{16} + \calL_{32}. \label{eq:final_loss}
\end{align}
The motivations for \eqref{eq:loss}-\eqref{eq:final_loss} are as follows:
\begin{itemize}
\item the choice of Huber loss $\rho(\cdot)$ yields robustness to ground truth outliers;
\item  \eqref{eq:loss_incr}
is  invariant to rotations which suits well  IMUs, whose gyro and
accelerometer measurements are
respectively invariant to rotations and yaw changes  \cite{scaramuzzaVisualInertial2019,zhangTutorial2018
}, i.e., left shifting $\bfR_n\leftarrow\delta \bfR\bfR_n$ and
$\hbfR_n\leftarrow\delta \bfR\hbfR_n$ with $\delta\bfR_n \in SO(3)$
leaves \eqref{eq:loss} unchanged;
\item the choice of \eqref{eq:final_loss} corresponds to error increments at $200/16\approx\SI{12}{Hz}$ and $200/32\approx\SI{6}{Hz}$, which is barely slower than ground truth. Setting too high a $j$, or in the extreme case  using a loss based on the overall  orientation error   $\bfR_n^T\hbfR_n$, would make the algorithm prone to overfitting, and hence makes the method too sensitive to specific trajectory patterns of training
data.
\end{itemize}

\renewcommand{\figurename}{Table}
\setcounter{figure}{0} 
\begin{figure*}
	\centering
	\scriptsize
	\begin{tabular}{c|c||c|c|c|c|c|c|c|c|c}
		\toprule
\multirow{2}{*}{dataset} & \multirow{2}{*}{sequence} &  VINS- &VINS-Mono & Open-
& Open-VINS &
   zero 
 & raw  & OriNet&  calibrated IMU & \textbf{proposed}\\
&  & Mono \cite{qinVINSMono2018}& (loop-closure)&VINS
\cite{genevaOpenVINS2019}& (\textbf{proposed})  &motion  &IMU  & \cite{esfahaniOriNet2020} &  (\textbf{proposed})
& \textbf{IMU} \\ \midrule
 
& MH 02 easy &
1.34/1.32&\textbf{0.57}/\textbf{0.50}& 1.11/1.05& 1.21/1.12& 44.4/43.7
&146/130&5.75/0.51&7.09/1.49&1.39/0.85\\ \cmidrule{2-11}

& MH 04 difficult&
1.44/1.40&\textbf{1.06}/1.00&1.60/1.16&1.40/0.89&42.3/41.9&
130/77.9&8.85/7.27 &5.64/2.53&1.40/\textbf{0.25} \\\cmidrule{2-11}

EuRoC&V1 01 easy&
0.97/0.90&\textbf{0.57}/\textbf{0.44}&0.80/0.67&0.80/0.67&114/76&
71.3/71.2&6.36/2.09&6.65/3.95&1.13/0.49\\\cmidrule{2-11}

\cite{burriEuRoC2016}&  V1 03 difficult&
4.72/4.68&4.06/4.00&2.32/2.27&\textbf{2.25}/2.20&81.4/80.5&119/84.9&14.7/11.5
&3.56/2.04&2.70/\textbf{0.96}\\ \cmidrule{2-11}

&V2 02 medium&
2.58/2.41&1.83/1.61&1.85/1.61&\textbf{1.81}/\textbf{1.57}&93.9/93.5&117/86&11.7/6.03&
4.63/2.30&3.85/2.25\\ \cmidrule{2-11}

&\textbf{average}&
2.21/2.14&1.62/1.52&1.55/1.37&\textbf{1.50}/1.30&66.1/66.1&125/89.0&9.46/5.48
&5.51/2.46&2.10/\textbf{0.96}\\ \midrule

& room 2&
\textbf{0.60}/\textbf{0.45} &0.69/0.50 &2.47/2.36&1.95/1.84&91.8/90.4&
118/88.1&--/--&10.6/10.5&1.31/1.18\\ \cmidrule{2-11}

TUM-VI&room 4&
0.76/0.63 &\textbf{0.66}/\textbf{0.51}& 0.97/0.88&0.93/0.83&107/103&
74.1/48.2&--/--&2.43/2.30&1.48/0.85\\ \cmidrule{2-11}

\cite{schubertTUM2018}&  room 6&
0.58/0.38 &\textbf{0.54}/\textbf{0.33}&0.63/0.51&0.60/0.51&138/131&
94.0/76.1&--/--&4.39/4.31&1.04/0.57\\ \cmidrule{2-11}

&\textbf{average}&
0.66/0.49 &\textbf{0.63}/\textbf{0.45}&1.33/1.25&1.12/1.05&112/108&
95.7/70.8&--/--&5.82/5.72&1.28/0.82\\
\bottomrule
\end{tabular}

\caption{Absolute Orientation
Error (AOE) in terms of \textbf{3D orientation/yaw}, in degree, on the \emph{test} sequences. We see   our approach competes with VIO (while entirely based on IMU signals) and
outperforms other inertial methods. Results from OriNet are  unavailable on the TUM-VI
dataset. \label{fig:aoe}}
\vspace*{-0.6cm}
\end{figure*}
\renewcommand{\figurename}{Fig.}
\setcounter{figure}{4} 

\subsection{Efficient Computation of \eqref{eq:loss_incr}-\eqref{eq:final_loss}}

 First, note that thanks to parallelization 
applying e.g. $\exp(\cdot)$, to one or parallelly to many   $\hat{\boomega}_n$ takes similar execution time   on a GPU (we found experimentally that one operation takes   \SI{5}{ms} whereas 10 million operations in parallel takes \SI{70}{ms}, that is, the time per operation drops to \SI{7}{ns}). We call an operation that is parallelly applied to many instances a \emph{batch operation}. That said, an apparent drawback of \eqref{eq:loss_incr}
is to require $j$ matrix multiplications, i.e. $j$ operations. However, first, we may compute ground truth
$\delta \bfR_{i,i+j}$ only once, store it, and then we  only need to compute $\delta
\hbfR_{i,i+j}$ multiple times. Second, by viewing \eqref{eq:loss_incr} as a
tree of matrix multiplications, see Figure \ref{fig:tree},
we reduce the computation to $\log_2(j)$ batch GPU operations only. We finally apply
sub-sampling and take one $i$ every $j$ timestamps to avoid counting multiple times the same increment. Training speed is thus increased by a factor
$32/\log_2(32) \approx 6$.

\subsection{Training with Data Augmentation}

Data augmentation is a way to significantly increase the diversity of data
available for training without actually collecting new data, to avoid overfitting. This may be done for the IMU model  of Section \ref{sec:model} by  adding Gaussian noise
$\boeta_n$, adding static bias $\bfb_n$, uncalibrating the IMU, and shifting the
orientation of the IMU in the accelerometer measurement. The two first points were
  noted in \cite{esfahaniOriNet2020}, whereas the two latter are to the
best of our knowledge novel.

Although each point may increase the diversity of data, we   found
they do not necessarily improve the results. We opted for   addition of 
a Gaussian noise (only),  during each training epoch, whose standard deviation is the half the standard deviation that
the dataset provides (\SI{0.01}{deg/s}).

\section{Experiments}\label{sec:experiments}
We evaluate the method in terms of 3D orientation and yaw estimates, as the latter are more critical regarding long-term odometry estimation \cite{delmericoBenchmark2018,scaramuzzaVisualInertial2019}.

\subsection{Dataset Descriptions}\label{sec:dataset}
We   divide data into training, validation, and
test sets,  defined as follows, see Chapter I.5.3 of \cite{goodfellowDeep2016}. We
optimize the neural network and calibration parameters on
the training set. Validation set intervenes when training is over, and 
provides a \emph{biased} evaluation, as  the validation set is
used for training (data are seen, although never used for ``learning").
The test set is the gold standard to provide an \emph{unbiased} evaluation. It is only
used once training (using the training and validation sets) is terminated. The datasets we use are as follows.

\subsubsection{EuRoC} the dataset \cite{burriEuRoC2016} contains image and inertial
data at \SI{200}{Hz} from a \emph{micro aerial vehicle} divided into 11 flight
trajectories of 2-3 minutes in two environments. The ADIS16448 IMU is
\emph{uncalibrated} and we note ground truth  from laser tracker and motion
capture system is accurately
time-synchronized with the IMU, although dynamic
motions deteriorate the measurement accuracy. As yet
noticed in \cite{genevaOpenVINS2019}, ground truth for the sequence \texttt{V1}
\texttt{01} \texttt{easy} needs to be
recomputed. The sequences \texttt{MH} and \texttt{V1} were acquired in two consecutive days, and the sequences \texttt{V2} more than three months latter. Thus the network should adapt to varying bias.

We define the train set as the first \SI{50}{s} of the six sequences \texttt{MH\{01,03,05\}},
\texttt{V1\{02\}}, \texttt{V2\{01,03\}}, the validation set as the remaining ending parts of these sequences, and we constitute the test set as the five remaining sequences. We show in Section \ref{sec:results} that using only 8 minutes of accurate data for training - the beginning and end of each
trajectory are the most accurately measured - is sufficient to obtain relevant results.

\subsubsection{TUM-VI}
the recent dataset \cite{schubertTUM2018} consists of visual-inertial sequences in different scenes from an \emph{hand-held} device. The cheap BMI160 IMU logs data at \SI{200}{Hz} and
was properly \emph{calibrated}. Ground truth is  accurately
time-synchronized with the IMU, although each sequence contains periodic instants of duration
\SI{0.2}{s} where ground truth is \emph{unavailable} as the acquisition platform
was hidden from the motion capture system, see Figure \ref{fig:gt2}. We take the
6 room sequences, which are the sequences having  longest ground truth (2-3
minutes each).

We define the train set as the first \SI{50}{s} of the sequences \texttt{room} \texttt{1}, \texttt{room} \texttt{3}, and \texttt{room} \texttt{5}, the validation set as
the remaining ending parts of these sequences, and we set the test set as the 3 other room sequences. This slipt corresponds to \num{45000} training data points (\num{90000} for
EuRoC) which is in the same order as the number of optimized parameters, \num{77052}, and
requires regularization techniques such as weight decay and dropout during training.

\begin{figure}
	\centering
	\ifthenelse{\boolean{silvere}}{
		\includegraphics{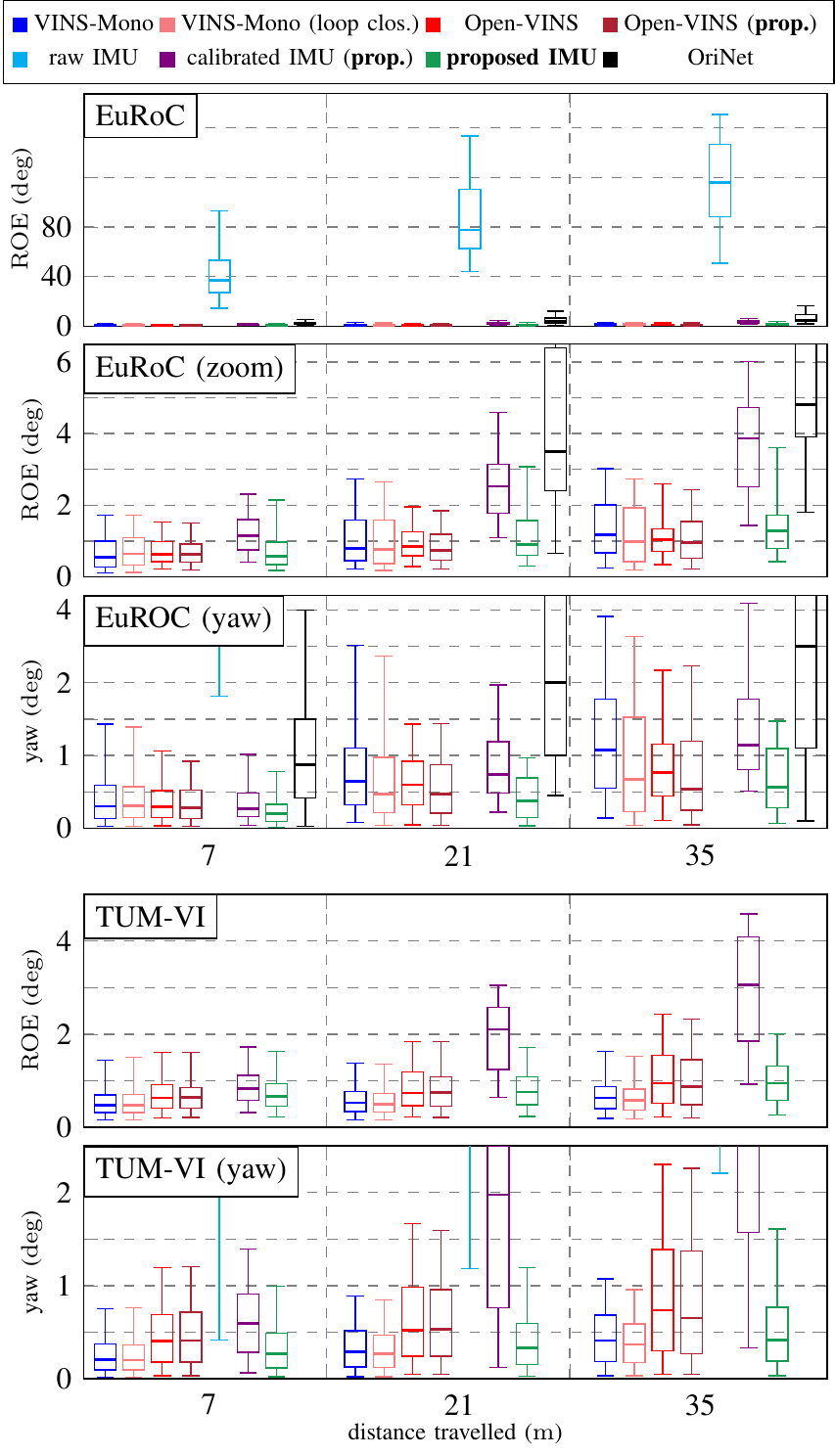}
	}{
		\include{boxplot}
	}
	\vspace*{-0.7cm}
	\caption{Relative Orientation Error (ROE) in terms of 3D orientation and yaw errors on the test sequences. Our method outperforms calibrated IMU and competes with VIO methods albeit based only on IMU signals. Raw IMU is way off and results from OriNet on TUM-VI are unavailable. \label{fig:roe}}
	\vspace*{-0.4cm}
  \end{figure}

\subsection{Method Implementation \& Training}\label{sec:train}
Our open-source  method is implemented based on PyTorch 1.5, where we configure the
training hyperparameters as follows. We set weight decay with parameter
0.1, and dropout
with 0.1 the probability of an element to be set equal to zero. Both techniques reduce
overfitting. 

We choose the ADAM optimizer \cite{kingmaAdam2014} with cosines warning restart scheduler \cite{loshchilovSGDR2016} where
learning rate is initialized at 0.01. We train for \num{1800} epochs, which is
is very fast as it takes less than 5 minutes for each dataset with a GTX 1080 GPU
device.

\begin{figure*}
	\centering
	\ifthenelse{\boolean{silvere}}{
		\includegraphics{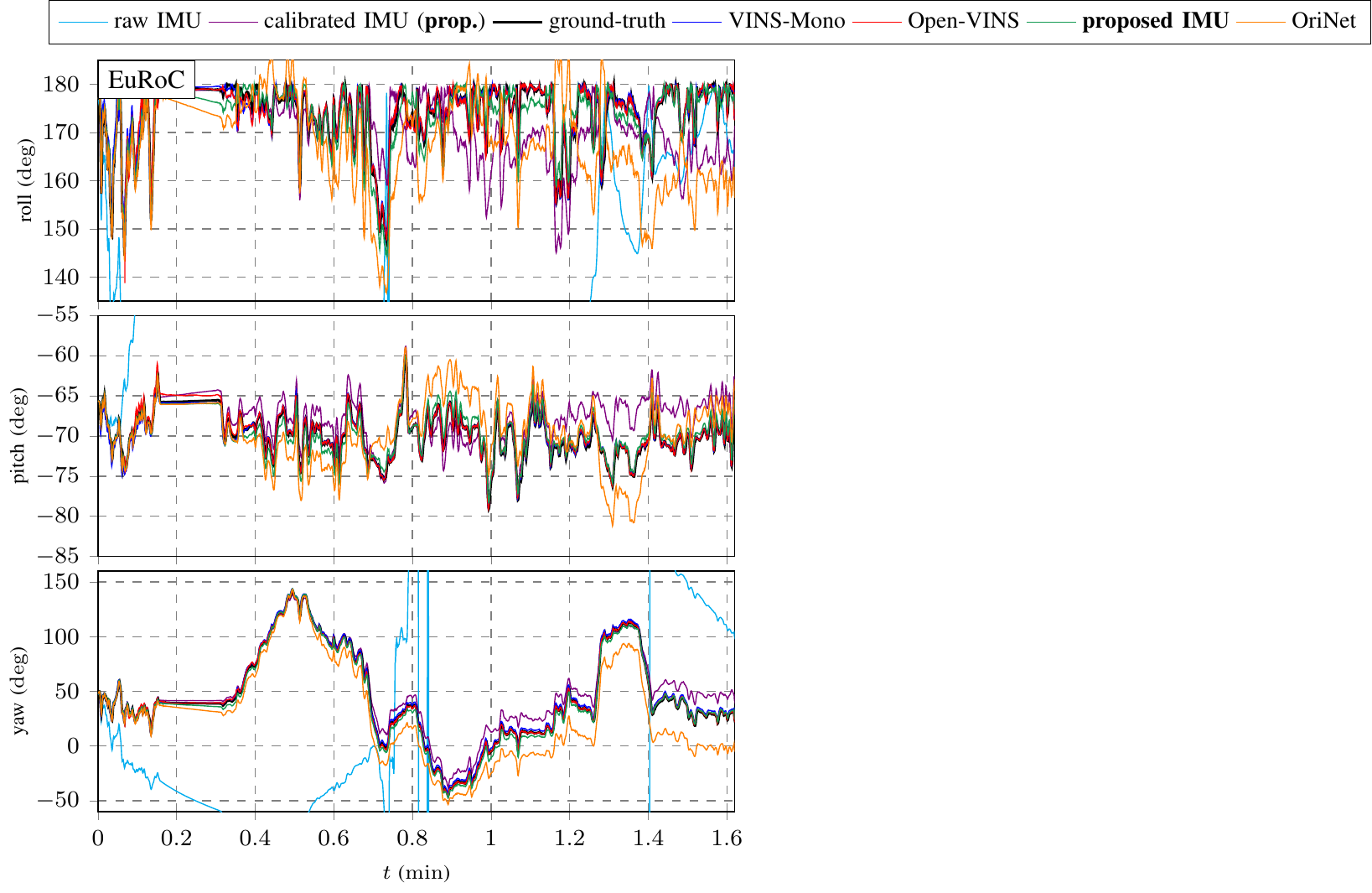}
		\hspace{-7.3cm}
		\includegraphics{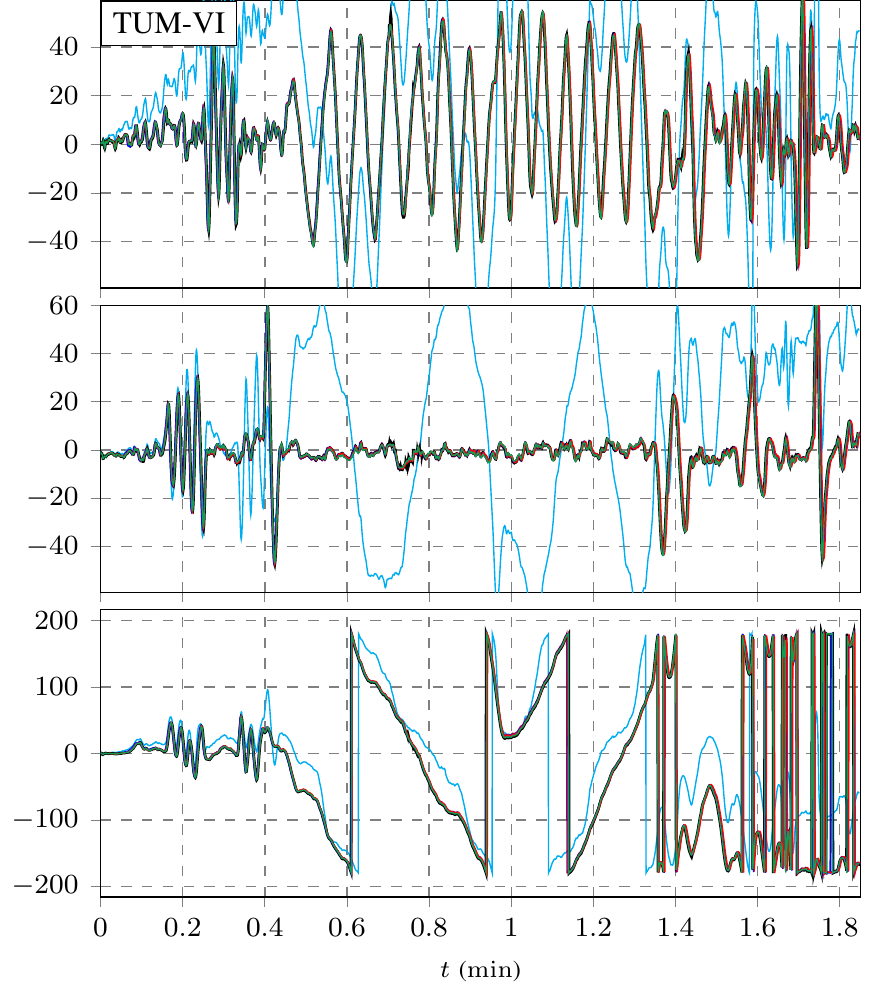}
	}{
		        

\begin{tikzpicture}
\begin{groupplot}[
    group style={
        group name=my plots,
        group size=1 by 3,
        xlabels at=edge bottom,
        xticklabels at=edge bottom,
        vertical sep=5pt,
    },
    footnotesize,
    width=9.3cm,
    height=4.5cm,
    xlabel={$t$ (\si{min})},
    xtick={0,0.2,...,1.4, 1.6},
    xlabel style={font=\scriptsize},
    ylabel style={font=\scriptsize},
    xtick style={font=\scriptsize},
    ytick style={font=\scriptsize},
    xmin=0, xmax=1.62,
    tickpos=left,
    ytick align=outside,
    xtick align=outside,
    ylabel shift = -0.15cm,
    ticks=both,
    grid=both,
]
\nextgroupplot[ylabel={roll (\si{deg})},
legend columns=7,
xlabel style={font=\scriptsize},
ylabel style={font=\scriptsize},
xtick style={font=\scriptsize},
ytick style={font=\scriptsize},
legend style ={font=\small},
legend entries={raw IMU, calibrated IMU (\textbf{prop.}), ground-truth, VINS-Mono, Open-VINS,
{\textbf{proposed IMU}}, OriNet},
title={\small EuRoC},every axis title/.style={below
right,at={(0,1)},draw=black,fill=white},
legend style={at={(2,1.25)},font=\footnotesize}, ymax=185, ymin=135]
\addplot[draw=cyan]  table[x=t,y=imu_roll] {figures/mh_04.txt};
\addplot[draw=violet]  table[x=t,y=net_roll] {figures/mh_04.txt};
\addplot[draw=black, thick]  table[x=t,y=roll] {figures/mh_04.txt};
\addplot[draw=blue]  table[x=t,y=net_roll] {figures/mh_04_vm.txt};
\addplot[draw=red]  table[x=t,y=net_roll] {figures/mh_04_ov.txt};
\addplot[draw=ForestGreen]  table[x=t,y=net_roll] {figures/mh_04_2.txt};
\addplot[draw=orange]  table[x=t,y=roll] {figures/orinet.txt};

\nextgroupplot[ylabel={pitch (\si{deg})}, 
xlabel style={font=\scriptsize},
ylabel style={font=\scriptsize},
xtick style={font=\scriptsize},
ytick style={font=\scriptsize},
ymax=-55, ymin=-85]
\addplot[draw=cyan]  table[x=t,y=imu_pitch] {figures/mh_04.txt};
\addplot[draw=violet]  table[x=t,y=net_pitch] {figures/mh_04.txt};
\addplot[draw=black, thick]  table[x=t,y=pitch] {figures/mh_04.txt};
\addplot[draw=blue]  table[x=t,y=net_pitch] {figures/mh_04_vm.txt};
\addplot[draw=red]  table[x=t,y=net_pitch] {figures/mh_04_ov.txt};
\addplot[draw=ForestGreen]  table[x=t,y=net_pitch] {figures/mh_04_2.txt};
\addplot[draw=orange]  table[x=t,y=pitch] {figures/orinet.txt};

\nextgroupplot[ylabel={yaw (\si{deg})}, 
xlabel style={font=\scriptsize},
ylabel style={font=\scriptsize},
xtick style={font=\scriptsize},
ytick style={font=\scriptsize},
ymax=160, ymin=-60]
\addplot[draw=cyan]  table[x=t,y=imu_yaw] {figures/mh_04.txt};
\addplot[draw=violet]  table[x=t,y=net_yaw] {figures/mh_04.txt};
\addplot[draw=black, thick]  table[x=t,y=yaw] {figures/mh_04.txt};
\addplot[draw=blue]  table[x=t,y=net_yaw] {figures/mh_04_vm.txt};
\addplot[draw=red]  table[x=t,y=net_yaw] {figures/mh_04_ov.txt};
\addplot[draw=ForestGreen]  table[x=t,y=net_yaw] {figures/mh_04_2.txt};
\addplot[draw=orange]  table[x=t,y=yaw] {figures/orinet.txt};
\end{groupplot}
\end{tikzpicture}
\hspace*{-7.9cm}
\begin{tikzpicture}
\begin{groupplot}[
    group style={
        group name=my plots,
        group size=1 by 3,
        xlabels at=edge bottom,
        xticklabels at=edge bottom,
        vertical sep=5pt
    },
    footnotesize,
    width=9.3cm,
    height=4.5cm,
    xlabel={$t$ (\si{min})},
    xlabel style={font=\scriptsize},
    ylabel style={font=\tiny},
    xtick style={font=\tiny},
    ytick style={font=\scriptsize},
    xmin=0, xmax=1.85,
    xtick={0,0.2,...,1.4,1.6,1.8},
    tickpos=left,
    ytick align=outside,
    xtick align=outside,
    ylabel shift = -0.2cm,
    ticks=both,
    grid=both,
]
\nextgroupplot[
xlabel style={font=\tiny},
ylabel style={font=\tiny},
xtick style={font=\tiny},
ytick style={font=\tiny},
title={TUM-VI},font=\small,every axis title/.style={below
right,at={(0,1)},draw=black,fill=white},
legend style={at={(1.0,1.45)},font=\footnotesize}, ymax=59, ymin=-59]

\addplot[draw=cyan]  table[x=t,y=net_roll] {figures/4_imu.txt};
\addplot[draw=violet]  table[x=t,y=net_roll] {figures/4_calib.txt};
\addplot[draw=black, thick]  table[x=t,y=net_roll] {figures/4_gt.txt};
\addplot[draw=blue]  table[x=t,y=net_roll] {figures/4_vm.txt};
\addplot[draw=red]  table[x=t,y=net_roll] {figures/4_ov.txt};
\addplot[draw=ForestGreen]  table[x=t,y=net_roll] {figures/4_net.txt};

\nextgroupplot[ymax=60, 
xlabel style={font=\scriptsize}, 
ylabel style={font=\scriptsize},
xtick style={font=\scriptsize},
ytick style={font=\scriptsize},
ymin=-59]
\addplot[draw=cyan]  table[x=t,y=net_pitch] {figures/4_imu.txt};
\addplot[draw=violet]  table[x=t,y=net_pitch] {figures/4_calib.txt};
\addplot[draw=black, thick]  table[x=t,y=net_pitch] {figures/4_gt.txt};
\addplot[draw=blue]  table[x=t,y=net_pitch] {figures/4_vm.txt};
\addplot[draw=red]  table[x=t,y=net_pitch] {figures/4_ov.txt};
\addplot[draw=ForestGreen]  table[x=t,y=net_pitch] {figures/4_net.txt};

\nextgroupplot[xlabel style={font=\scriptsize},
ylabel style={font=\scriptsize},
xtick style={font=\scriptsize},
ytick style={font=\scriptsize}]
\addplot[draw=cyan]  table[x=t,y=net_yaw] {figures/4_imu.txt};
\addplot[draw=violet]  table[x=t,y=net_yaw] {figures/4_calib.txt};
\addplot[draw=black, thick]  table[x=t,y=net_yaw] {figures/4_gt.txt};
\addplot[draw=blue]  table[x=t,y=net_yaw] {figures/4_vm.txt};
\addplot[draw=red]  table[x=t,y=net_yaw] {figures/4_ov.txt};
\addplot[draw=ForestGreen]  table[x=t,y=net_yaw] {figures/4_net.txt};
\end{groupplot}
\end{tikzpicture}
	}
\vspace*{-0.8cm}
\caption{Orientation estimates on the test sequence \texttt{MH 04 difficult} of
\cite{burriEuRoC2016} (left), and \texttt{room} \texttt{4} of \cite{schubertTUM2018}
(right). Our method removes errors of the calibrated IMU and competes with VIO
algorithms.\label{fig:rpy}}
\vspace*{-0.4cm}
\end{figure*}

\subsection{Compared Methods}\label{sec:compared}
We compare   a set of methods based on camera and/or IMU.

\subsubsection{Methods Based on the IMU Only} we compare the following approaches:
\begin{itemize}
\item \textbf{raw IMU}, that is an uncalibrated IMU. It refers also to the proposed method once initialized but not trained;
\item  \textbf{OriNet} \cite{esfahaniOriNet2020}, which is based on \emph{recurrent
neural networks}, and whose training and test sets correspond to ours;
\item \textbf{calibrated IMU}, that is, our method where the 12 parameters
$\hat{\bfC}_{\boomega}$ and $\tilde{\boomega}_n$ are \emph{constant}, nonzero, and
optimized. This can be seen by enforcing to zero the network inputs, i.e. setting the IMU gyros and accelerometers in \eqref{eq:tilde} as $\bfu_n=\bfzero_6$, both during training and evaluation;
	\item \textbf{proposed IMU}, which is our learning based method described in Section \ref{sec:method}.
\end{itemize}

\subsubsection{Methods Based on Camera and the IMU} we run each of the following method with the same 
setting, ten times to then average results, and on a Dell Precision Tower 7910 workstation desktop, i.e., without
deterioration due to computational limitations
\cite{delmericoBenchmark2018}. We compare:
\begin{itemize}
\item  \textbf{VINS-Mono} \cite{qinVINSMono2018}, a monocular VIO framework  with notable performances on the EuRoC benchmark
\cite{delmericoBenchmark2018};
\item \textbf{VINS-Mono (loop closure)}, which is the original VINS-Mono \cite{qinVINSMono2018}
reinforced with loop-closure ability;
\item \textbf{Open-VINS} \cite{genevaOpenVINS2019}, a versatile filter-based
visual-inertial estimator for which we choose the  stereo configuration, and that
is top-ranked on the drone dataset of \cite{delmericoAre2019};
\item \textbf{Open-VINS (proposed)}, which is Open-VINS of \cite{genevaOpenVINS2019} but where  
gyro inputs are the proposed corrected gyro measurements \eqref{eq:homega} outputted by our method (trained on sequences that are of course different from those used for evaluation).
\end{itemize}

\subsection{Evaluation Metrics}\label{sec:metrics}
We evaluate the above methods using the following metrics that we compute  with the
 toolbox of \cite{genevaOpenVINS2019}.

\subsubsection{Absolute Orientation Error (AOE)} which computes the mean square error
between the ground truth and estimates for a given sequence as
\begin{align}
	\mathrm{AOE}=\sqrt{\sum_{n=1}^{M} \frac{1}{M} \|  \log \left(\bfR_{n}^T \hbfR_{n}\right) \|_2^2}, \label{eq:aoe}
	\end{align}
with $M$ the sequence length, $\log(\cdot)$ the $SO(3)$ logarithm map, and where
the estimated trajectory  has been aligned on the ground truth
at the first instant $n=0$.

\subsubsection{Relative Orientation Error (ROE)} which is computed as \cite{zhangTutorial2018}
\begin{align}
\mathrm{ROE}= \|\log \left(\delta \bfR_{n,g(n)}^T \delta \hbfR_{n,g(n)}\right)\|_2, \label{eq:roe}
\end{align}
for each pair of timestamps $\left(n, g(n)\right)$ representing an IMU displacement of
7, 21 or 35 meters. Collecting  the error \eqref{eq:roe} for  all the pairs of
sub-trajectories  generates a collection of errors where informative statistics such as the
median and percentiles are calculated. As \cite{zhangTutorial2018,delmericoBenchmark2018,genevaOpenVINS2019}, we
strongly recommend ROE for comparing odometry
estimation methods since 
AOE is  highly sensitive to the time when the
estimation error occurs.  We finally consider slight variants of
\eqref{eq:aoe}-\eqref{eq:roe} when considering \emph{yaw} (only)  errors, and
note that errors of visual methods generally scale with distance travelled whereas
errors of inertial only methods scales with time. We provide in the present
paper errors w.r.t.
distance travelled to favor comparison with benchmarks such as
\cite{delmericoBenchmark2018}, and same conclusions hold when computing ROE as function
of different times.

\subsection{Results}\label{sec:results}

\begin{figure}
	\centering
	\ifthenelse{\boolean{silvere}}{
		\includegraphics{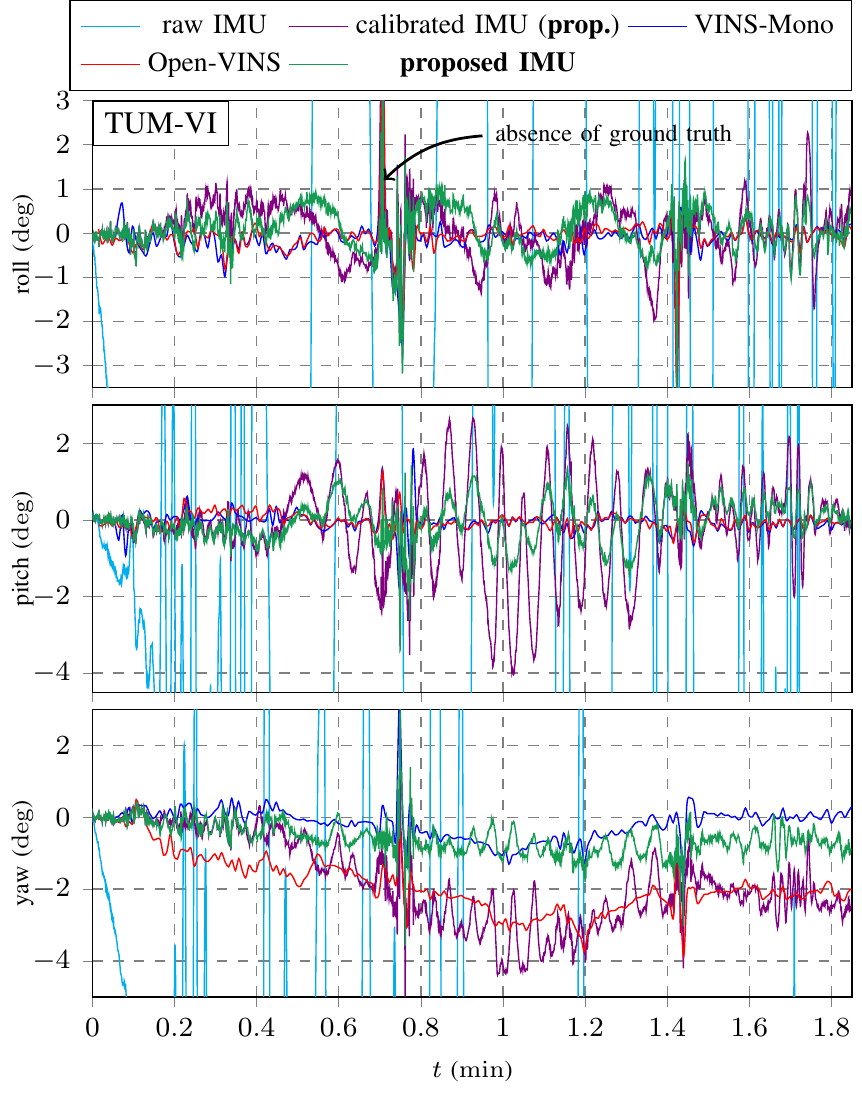}
	}{
		\begin{tikzpicture}
\begin{groupplot}[
    group style={
        group name=my plots,
        group size=1 by 3,
        xlabels at=edge bottom,
        xticklabels at=edge bottom,
        vertical sep=5pt
    },
    footnotesize,
    width=9.3cm,
    height=4.5cm,
    xlabel={$t$ (\si{min})},
    xlabel style={font=\scriptsize},
    ylabel style={font=\scriptsize},
    xtick style={font=\scriptsize},
    ytick style={font=\scriptsize},
    xmin=0, xmax=1.85,
    xtick={0,0.2,...,1.4,1.6,1.8},
    tickpos=left,
    ytick align=outside,
    xtick align=outside,
    ylabel shift = -0.25cm,
    ticks=both,
    grid=both,
]
\nextgroupplot[ylabel={roll (\si{deg})},
legend columns=3,
legend style ={font=\small},
legend entries={raw IMU, calibrated IMU (\textbf{prop.}), VINS-Mono , Open-VINS,{\textbf{proposed IMU}}},
title={TUM-VI},font=\small,every axis title/.style={below
right,at={(0,1)},draw=black,fill=white},
legend style={at={(1,1.35)},font=\footnotesize}, ymax=3, ymin=-3.5]
\addplot[draw=cyan,smooth]  table[x=t,y=roll] {figures/4_imu_err.txt};
\addplot[draw=violet,smooth]  table[x=t,y=roll] {figures/4_calib_err.txt};
\addplot[draw=blue,smooth]  table[x=t,y=roll] {figures/4_vm_err.txt};
\addplot[draw=red,smooth]  table[x=t,y=roll] {figures/4_ov_err.txt};
\addplot[draw=ForestGreen,smooth]  table[x=t,y=roll] {figures/4_prop_err.txt};
\draw[black, thick, ->] (0.95,2.2) node[right,xshift=0.0cm] {\scriptsize absence
of ground truth} to[bend right = 20] (0.71,1.2);

\nextgroupplot[ylabel={pitch (\si{deg})}, ymax=3, ymin=-4.5]
\addplot[draw=cyan,smooth]  table[x=t,y=pitch] {figures/4_imu_err.txt};
\addplot[draw=violet,smooth]  table[x=t,y=pitch] {figures/4_calib_err.txt};
\addplot[draw=blue,smooth]  table[x=t,y=pitch] {figures/4_vm_err.txt};
\addplot[draw=red,smooth]  table[x=t,y=pitch] {figures/4_ov_err.txt};
\addplot[draw=ForestGreen,smooth]  table[x=t,y=pitch] {figures/4_prop_err.txt};

\nextgroupplot[ylabel={yaw (\si{deg})}, ymax=3, ymin=-5]
\addplot[draw=cyan,smooth]  table[x=t,y=yaw] {figures/4_imu_err.txt};
\addplot[draw=violet,smooth]  table[x=t,y=yaw] {figures/4_calib_err.txt};
\addplot[draw=blue,smooth]  table[x=t,y=yaw] {figures/4_vm_err.txt};
\addplot[draw=red,smooth]  table[x=t,y=yaw] {figures/4_ov_err.txt};
\addplot[draw=ForestGreen,smooth]  table[x=t,y=yaw] {figures/4_prop_err.txt};

\end{groupplot}
\end{tikzpicture}
	}
	\vspace*{-0.9cm}
	\caption{Orientation \emph{errors} on the sequence \texttt{room} \texttt{4} of \cite{schubertTUM2018}. Our method removes errors of the calibrated IMU and competes with VIO algorithms.\label{fig:rpy_err}}
	\vspace*{-0.6cm}
\end{figure}

Results are given  in term of AOE and ROE respectively in Table \ref{fig:aoe}
and Figure \ref{fig:roe}. Figure \ref{fig:rpy} illustrates roll, pitch and
yaw estimates for a test sequence of each dataset, and Figure \ref{fig:rpy_err}
shows orientation \emph{errors}. We note that:

\subsubsection{Uncalibrated IMU is Unreliable} raw IMU estimates  deviate from
ground truth in less than \SI{10}{s}, see Figure \ref{fig:rpy}. 

\subsubsection{Calibrated IMU Outperforms OriNet} only calibrating an IMU (via our optimization method) leads
to surprisingly accurate results, see e.g.,  Figure
\ref{fig:rpy} (right) where it is difficult to distinguish it  from ground truth.
This evidences cheap sensors can provide very accurate  information once they are
correctly calibrated.

\subsubsection{The Proposed Method Outperforms Inertial Methods}   OriNet \cite{esfahaniOriNet2020} is outperformed. Moreover, our method improves   accurate calibrated IMU
by a factor
2 to 4. Our approach notably obtains as low as a median error of \SI{1.34}{\deg/min} and
\SI{0.68}{\deg/min} on respectively EuRoC and TUM-VI datasets.

\subsubsection{The Proposed Method Competes with VIO}
our IMU only method is accurate even on the high motion dynamics present
in both datasets, see Figure \ref{fig:rpy}, and competes with VINS-Mono and Open-VINS, although
trained with only a  few minutes of data.

Finally, as the performance of  each method  depends on the dataset and the algorithm setting, see
Figure \ref{fig:roe}, it is difficult  to
conclude  which VIO algorithm is the best.

\subsection{Further Results and Comments}

We provide a few  more comments, supported by further experimental results.

\subsubsection{Small Corrections Might Lead   to Large Improvement} 
\begin{figure}
	\centering
	\ifthenelse{\boolean{silvere}}{
		\includegraphics{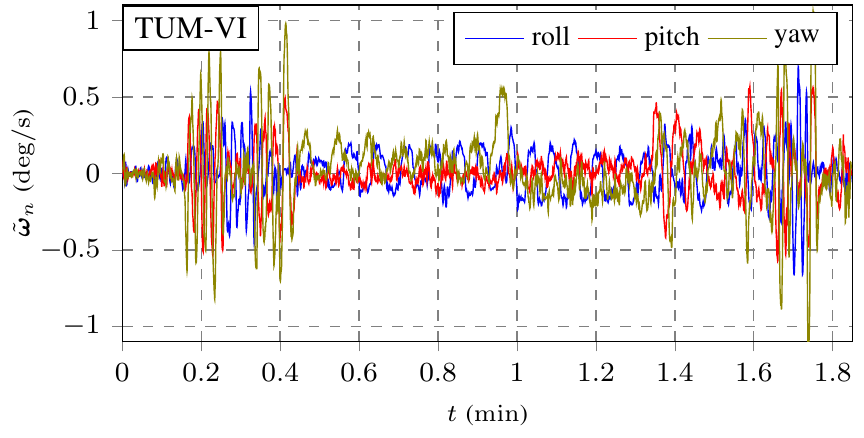}
	}{
		\begin{tikzpicture}
\begin{axis}[
	footnotesize,
	width=9cm,
	height=5cm,
	xlabel={$t$ (\si{min})},
	ylabel={$\tilde{\boomega}_n$ (\si{deg/s})},
	xlabel style={font=\scriptsize},
	ylabel style={font=\scriptsize},
	xtick style={font=\scriptsize},
	ytick style={font=\scriptsize},
	xmin=0, xmax=1.85,
	ymin = -1.1,
	ymax = 1.1,
	xtick={0,0.2,...,1.6,1.8},
	tickpos=left,
	ytick align=outside,
	xtick align=outside,
	ylabel shift = -0.2cm,
	ticks=both,
	grid=both,
	legend columns=3,
	legend style ={font=\footnotesize},
	legend entries={roll, pitch, yaw},
	title={TUM-VI},font=\small,every axis title/.style={below
	right,at={(0,1)},draw=black,fill=white}]

\addplot[draw=blue,smooth]  table[x=t,y=roll] {figures/4_gyro_correction.txt};
\addplot[draw=red,smooth]  table[x=t,y=pitch] {figures/4_gyro_correction.txt};
\addplot[draw=olive,smooth]  table[x=t,y=yaw] {figures/4_gyro_correction.txt};
\end{axis}
\end{tikzpicture}
	}
	\vspace*{-1cm}
\caption{Gyro correction $\tilde{\boomega}_n$ on the sequence \texttt{room} \texttt{4} of
\cite{schubertTUM2018}. We see we manage to divide  orientation errors by a factor at least 2 w.r.t. calibrated
IMU applying
corrections whose amplitude is as low as 
\SI{1}{deg/s} (max).\label{fig:correction}}
\vspace*{-0.6cm}
\end{figure}

the calibrated and
corrected gyro signals are visually undistinguishable: 
differences between them rely in 
corrections $\tilde{\boomega}_n$ of few \si{deg/s}, as shown in Figure
\ref{fig:correction}. However, they bring drastic improvement in the estimates. This confirms the interest of   leveraging neural networks for  
  model correction \eqref{eq:imu}-\eqref{eq:calib}.

\subsubsection{The Proposed Method is Well Suited to Yaw Estimation} according to Table \ref{fig:aoe} and Figure \ref{fig:roe}, we see yaw estimates are particularly accurate. Indeed, VIO methods are able to recover at any
time roll and pitch thanks to accelerometers, but the yaw   estimates drift with time.
In contrast our dead-reckoning method never has access to information allowing to  recover  roll and pitch during testing, and nor does it use  ``future''
information such as VINS-Mono with loop-closure ability. We finally
note that accurate yaw estimates could be fruitful for yaw-independent VIO methods such as
\cite{svachaInertial2019}.

\subsubsection{Correcting Gyro Slightly Improves Open-VINS
\cite{genevaOpenVINS2019}} both methods based on Open-VINS perform  similarly, which is not surprising  as camera alone already  provides
 accurate orientation estimates and the gyro  assists stereo cameras. 

\subsubsection{Our Method Requires few Computational Ressources}
each VIO method performs here at its  best while resorting to high computational requirements,
and we expect our method - once trained - is very  attractive when running onboard with limited resources. Note that, the proposed method  performs e.g. 3 times
better in terms of yaw estimates than  a slightly restricted VINS-Mono, see Figure 3 of
\cite{delmericoBenchmark2018}.

\section{Discussion}\label{sec:discuss}

We now provide the community with feedback regarding the method and its implementation. Notably, we emphasize a few points that seem key to a successful implementation when working with a low-cost high frequency IMU. 

\subsection{Key Points Regarding the Dataset}
One should be careful regarding the quality of data, especially when IMU is sampled at high-frequency. This concerns:
\subsubsection{IMU Signal} the IMU signal acquisition should be correct with constant sampling time.
\subsubsection{Ground Truth Pose Accuracy} we note that the EuRoC ground truth
accuracy is better at the beginning of the trajectory. As such,   training with only
this part of data (the first \SI{50}{s} of the training sequences) is sufficient (and best) to succeed.
\subsubsection{Ground Truth Time-Alignement} the time alignment between
ground truth and IMU is  significant for success,
otherwise the method is prone to learn a time delay.

We admit that our approach requires a proper dataset, which is what constitutes its main \emph{limitation}.

\subsection{Key Points Regarding the Neural Network}

 Our conclusions about the neural network  are as follows.
\subsubsection{Activation Function} the GELU and other smooth activation functions
\cite{ramachandranSearching2018},  such as   ELU, perform well, whereas ReLU based
network is more prone to
overfit. We believe ReLU activation function favors sharp corrections which does not make sense when dealing with 
physical signals.

\subsubsection{Neural Network Hyperparameters} increasing the depth, channel and/or kernel sizes of the network, see Figure \ref{fig:cnn}, does not systematically lead to
better results. We tuned these hyperparameters with random search, although more
sophisticated methods such as \cite{liHyperband2017} exist.

\subsubsection{Normalization Layer}  batchnorm layer improves both training speed and accuracy \cite{ioffeBatch2015}, and is highly recommended.

\subsubsection{Accelerometers are Relevant} we test our approach without accelerometer, a.k.a. ablation study experiment, where the neural network obtains 50 \% more errors than proposed IMU in term of ROE. It indicates how accelerometer are useful, see an instinctive reason in Section \ref{sec:cnn}.

\subsubsection{Reccurent Neural Network obtains Higher Errors}  we also compare our same approach where LSTM replaces the dilated convolutions in the network structure. This type of neural network obtains around 40 \% more errors than proposed IMU in term of ROE, while requiring also more time to be trained.

\subsection{Key Points Regarding Training}
As in any machine learning application, the neural network architecture is a key
component among others \cite{goodfellowDeep2016}. Our comments regarding training are as follows:
\subsubsection{Optimizer} the ADAM optimizer \cite{kingmaAdam2014} performs  well.
\subsubsection{Learning Rate Scheduler} adopting a learning rate policy with cosinus warning restart
\cite{loshchilovSGDR2016} leads to substantial improvement and helps to find a
correct learning rate.
\subsubsection{Regularization} dropout and weight decay hyperparameters are crucial to avoid overfitting.
Each has a range of ideal values which is quickly tuned with basic grid-search.

\subsection{Remaining Key Points}
We finally outline two points that we
consider useful to the practitioner: 
\subsubsection{Orientation Implementation} we did not find any difference between rotation matrix or quaternion loss function implementation once
numerical issues are solved, e.g., by enforcing quaternion unit norm.
Both implementations result in
similar accuracy performance and execution time.
\subsubsection{Generalization and Transfert Learning}  it may prove  useful to assess to what extent a learning method is generalizable. The extension of the method, trained on one
dataset, to another device or to the same
device on another platform is considered as challenging, though,  and left for future work.

\section{Conclusion}\label{sec:con}
This paper proposes a deep-learning method for denoising IMU gyroscopes and obtains
remarkable accurate attitude estimates with only a low-cost IMU, that outperforms state-of-the-art  \cite{esfahaniOriNet2020}. The core of
the approach is based on
a careful design and feature selection  of a dilated convolutional network, and an appropriate
loss function leveraged for training on orientation increment at the ground truth
frequency. This leads to a method robust to overfitting, efficient and fast to
train, which serves as   offline IMU calibration and may enhance it. As a remarkable byproduct, the
method  competes with state-of-the-art visual-inertial methods in term of attitude estimates on a drone and hand-held
device datasets, where we simply integrate noise-free gyro measurements.

We believe the present paper offers new perspectives for (visual-)inertial
learning methods. Future work  will address new challenges in three
directions: learning from multiple IMUs (the current method is reserved for one IMU only which serves for training and testing); learning from moderately accurate
ground truth that can be output of visual-inertial localization systems; 
and  denoising accelerometers based on relative increments from preintegration
theory \cite{forsterOnManifold2017,barrauMathematical2020}.

\section*{Acknowledgements}
The authors wish to thank Jeffrey Delmerico and Mahdi Esfahani for sharing respectively the results of the VIO
benchmark \cite{delmericoBenchmark2018} and the results of OriNet \cite{esfahaniOriNet2020}.

\bibliographystyle{IEEEtran}
\bibliography{biblio}
\end{document}